\documentclass{article}

\PassOptionsToPackage{table}{xcolor}

\PassOptionsToPackage{numbers,compress}{natbib}

\usepackage[preprint]{neurips_2026}

\usepackage[utf8]{inputenc}
\usepackage[T1]{fontenc}

\usepackage{url}
\usepackage{booktabs}
\usepackage{amsfonts}
\usepackage{amssymb}
\usepackage{nicefrac}
\usepackage{microtype}

\usepackage{xcolor}
\usepackage{graphicx}
\usepackage{caption}
\usepackage{array}
\usepackage{multirow}
\usepackage{makecell}
\usepackage{siunitx}
\usepackage{placeins}
\usepackage{tcolorbox}
\usepackage{enumitem}
\usepackage{hyperref}

\definecolor{rowgray}{RGB}{220,220,220}
\definecolor{gaingreen}{RGB}{0,150,0}

\definecolor{headerbg}{RGB}{247,244,248}     % very light mauve
\definecolor{paradigmbg}{RGB}{249,245,238}   % soft warm beige
\definecolor{closedbg}{RGB}{241,246,252}     % soft blue
\definecolor{unifiedbg}{RGB}{242,248,241}    % soft green
\definecolor{oursbg}{RGB}{255,244,214}       % soft highlight yellow
\definecolor{oursgray}{RGB}{215,215,215}
\definecolor{subgray}{RGB}{245,245,245}
\definecolor{mygreen}{RGB}{34,139,34}

\newcolumntype{C}{>{\hspace{0pt}}c<{\hspace{3.8pt}}}

\title{UniTriGen: Unified Triplet Generation of Aligned Visible-Infrared-Label for Few-Shot RGB-T Semantic Segmentation}

\author{%
\small
\begin{tabular}{@{}c@{}}
\textbf{Ping Zhou}\textsuperscript{1} \quad
\textbf{Haoyu Wang}\textsuperscript{1} \quad
\textbf{Mengmeng Zheng}\textsuperscript{1} \quad
\textbf{Lei Zhang}\textsuperscript{1,*} \\
\textbf{Wei Wei}\textsuperscript{1} \quad
\textbf{Chen Ding}\textsuperscript{2} \quad
\textbf{Fei Zhou}\textsuperscript{3} \\[2pt]
{\normalfont\mdseries \textsuperscript{1}School of Computer Science, Northwestern Polytechnical University} \\
{\normalfont\mdseries \textsuperscript{2}School of Computer Science \& Technology, Xi'an University of Posts \& Telecommunications} \\
{\normalfont\mdseries \textsuperscript{3}MMLab, The Chinese University of Hong Kong} \\[1pt]
{\normalfont\mdseries \texttt{zhou\_ping@mail.nwpu.edu.cn} \quad
\texttt{nwpuzhanglei@nwpu.edu.cn}} \\
{\normalfont\mdseries \textsuperscript{*}Corresponding authors.}
\end{tabular}%
}

\begin{document}

\maketitle

\begin{abstract}

RGB-T semantic segmentation requires strictly aligned VIS-IR-Label triplets; however, such aligned triplet data are often scarce in real-world scenarios. Existing generative augmentation methods usually adopt cascaded generation paradigms, decomposing joint triplet generation into local conditional processes. As a result, consistency among VIS, IR, and Label in spatial structure, semantic content, and cross-modal details cannot be reliably maintained. To address this issue, we propose UniTriGen, a unified triplet generation framework that directly generates spatially aligned, semantically consistent, and modality complementary VIS-IR-Label triplets under the guidance of text prompts. UniTriGen first introduces a unified triplet generation mechanism, where VIS, IR, and Label are jointly encoded into a shared latent space and modeled with a diffusion process to enforce global cross-modal consistency. Lightweight modality-specific residual adapters are further integrated into this mechanism to accommodate modality-specific imaging characteristics and output formats. To mitigate generation bias caused by imbalanced scene and class distributions in limited paired triplets, UniTriGen also employs a scene-balanced and class-aware few-shot sampling strategy, which induces a more balanced sampling distribution and enhances the scene and class diversity of generated triplets. Experiments show that UniTriGen generates high-quality aligned triplets from limited real paired data, thereby achieving consistent performance improvements across various RGB-T semantic segmentation models.

% RGB-T semantic segmentation requires strictly aligned VIS-IR-Label triplets, yet real, aligned triplets are often scarce. Existing generative augmentation methods usually rely on cascaded pipelines, which decompose joint triplet generation into local conditional processes and fail to reliably maintain cross-modal consistency. To address this issue, we propose UniTriGen, a unified text-driven framework for VIS-IR-Label triplet generation. UniTriGen introduces a unified triplet generation mechanism that jointly encodes VIS, IR, and Label into a shared latent space, models their concatenated latent representation with a diffusion process, and incorporates lightweight modality-specific residual adapters to preserve both cross-modal consistency and modality-specific fidelity. To reduce generation bias under limited paired triplet supervision, a scene-balanced and class-aware few-shot sampling strategy is further designed to induce a more balanced scene-class sampling distribution and improve the diversity of generated triplets. Experiments show that UniTriGen generates high-quality aligned triplets from limited real paired data and consistently improves various RGB-T semantic segmentation models.

\end{abstract}

\section{Introduction}
For RGB-T semantic segmentation, effective training samples should consist of strictly aligned same-scene Visible-Infrared-Label (VIS-IR-Label) triplets. The three modalities must share consistent scene layouts, object locations, boundaries, and semantic classes, so that visible appearance, infrared thermal responses, and pixel-level semantic supervision correspond within a common spatial coordinate frame. However, due to the practical challenges of multi-sensor acquisition, accurate cross-modal registration, and pixel-level semantic annotation, real RGB-T datasets typically provide only a small number of aligned and annotated triplets. Consequently, learning with only a limited number of paired triplets as training data becomes a more realistic yet more challenging setting.

A natural solution to this limitation is to use generative models to synthesize additional training triplets. However, this task is fundamentally different from generating each modality independently. Since VIS, IR, and Label are heterogeneous dense representations of the same scene, the generated triplets must preserve shared spatial structures, consistent semantic regions, and complementary cross-modal information. Therefore, the key challenge is not merely to generate more samples, but to learn the joint generative distribution of VIS, IR, and Label under the same scene from limited paired triplets.

\begin{figure}[htbp]
    \centering
    \vspace{-2pt}
    \includegraphics[width=\textwidth]{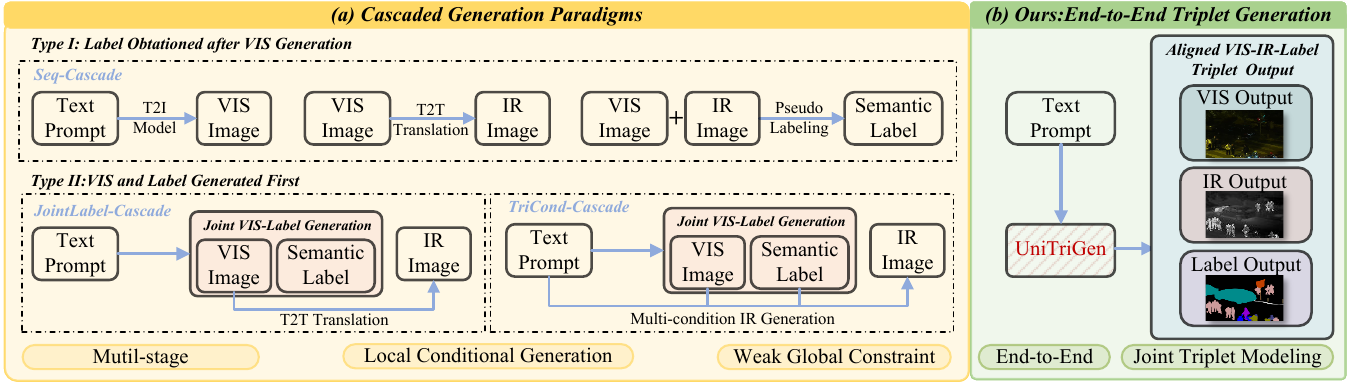}
    % \vspace{-10pt}
    \caption{Comparison between existing paradigms and the proposed UniTriGen. (a) Cascaded Generation Paradigms that progressively assemble the target triplet through intermediate outputs. (b) Our UniTriGen directly generates spatially aligned, semantically consistent, and modality complementary VIS-IR-Label triplets from text prompts in an end-to-end manner.}
    \vspace{-2pt}
    \label{figure1:motivation}
\end{figure}

To the best of our knowledge, text-driven generation of aligned VIS-IR-Label triplets has not yet been directly explored. Although existing generative paradigms can be adapted as alternatives, most follow cascaded generation paradigms. As shown in Figure \ref{figure1:motivation}(a), Seq-Cascade first generates visible images, then translates them into infrared images, and predicts the semantic labels. JointLabel-Cascade and TriCond-Cascade first generate image-label pairs and then synthesize the missing infrared images. Although these methods can produce triplets in form, they essentially decompose the target joint distribution into multiple local conditional generation processes. Specifically, each stage generates the next modality conditioned solely on the output of the preceding stage. Unlike a unified generation process that jointly constrains VIS, IR, and Label, this sequential design may propagate inconsistencies across stages. As a result, even if the output of each individual stage appears plausible, the final triplet may still suffer from layout shifts, boundary inconsistencies, semantic region mismatches, and cross-modal detail conflicts. In other words, cascaded paradigms are limited not only by error propagation and amplification along the generation chain, but also, more fundamentally, by their inability to impose global and unified cross-modal consistency constraints on the three modalities.

To address the above challenges, as shown in Figure~\ref{figure1:motivation}(b), we propose \textbf{UniTriGen}, a \textbf{Uni}fied \textbf{Tri}plet \textbf{Gen}eration framework that directly generates spatially aligned, semantically consistent, and modality-complementary VIS-IR-Label triplets under the guidance of text prompts. Specifically, a unified triplet generation mechanism is first designed   (Section~\ref{sec:unified_triplet_generation}), in which VIS images, IR images, and semantic labels are jointly encoded into a shared latent space. The resulting concatenated triplet latent representation is then modeled by a diffusion model, thereby explicitly enforcing cross-modal consistency among the three modalities. To account for the substantial differences among VIS, IR, and Label in imaging characteristics and decoding objectives, lightweight modality-specific residual adapters are further introduced during triplet decoding. These adapters improve modality-specific fidelity while preserving cross-modal consistency. In addition, when learning from only a limited number of paired triplets, the training data often exhibit highly imbalanced distributions of scenes and classes. This imbalance can bias the generative model toward frequent scenes and common classes, making it difficult to generate low-frequency scenes and rare classes. To mitigate this generation bias, a scene-balanced and class-aware few-shot sampling strategy is proposed (Section~\ref{sec:scene_balanced_sampling}). By constructing structured scene-class prompts, introducing explicit scene encoding, and applying hierarchical resampling, this strategy induces more balanced sampling distributions over scenes and classes. Consequently, the model’s dependence on frequent scenes and common classes is reduced, and the scene and class diversity of generated triplets is enhanced. With these designs, UniTriGen can generate many high-quality aligned VIS-IR-Label triplets using only a small number of real paired triplets, thereby improving the training of downstream RGB-T semantic segmentation models. Our main contributions are summarized as follows:

\begin{itemize}
    \item We propose UniTriGen, a text-to-VIS-IR-Label aligned triplet generation framework. The unified triplet generation mechanism jointly models VIS, IR, and Label in a shared latent space and employs lightweight modality-specific residual adapters for triplet decoding, producing spatially aligned, semantically consistent, and modality complementary triplets with improved modality fidelity and cross-modal consistency.

    % \item We design a scene-balanced and class-aware few-shot sampling strategy to mitigate generation bias from imbalanced scene and class distributions by constructing a more balanced sampling distribution, thereby enhancing the scene and class diversity of generated triplets.
    \item We design a scene-balanced and class-aware few-shot sampling strategy that integrates structured scene-class prompts, explicit scene encoding, and hierarchical resampling to reduce bias toward frequent scenes and common classes, thereby enhancing the diversity of generated triplets across scenes and classes.

    \item We conduct extensive experiments on SemanticRT and PST900. Results show that the high-quality synthetic triplets generated by UniTriGen effectively improve existing RGB-T semantic segmentation models and exhibit strong transferability across downstream architectures and data settings.
\end{itemize}

\section{Related Works}
\subsection{Dataset Generation for Semantic Segmentation}

Data generation for semantic segmentation\citep{TextSSR}\citep{Text2Earth}\citep{wu2023datasetdm}\citep{Re1_Dataset_diffusion}\citep{zhao2025pseudo}\citep{wang2026jodiffusion} has recently attracted increasing attention as a task-oriented data synthesis strategy. Unlike generic image synthesis\citep{SD}\citep{controlnet}\citep{ruiz2023dreambooth}\citep{labs2025flux}\citep{podellsdxl}, this line of work aims to produce training samples that improve pixel-level recognition rather than merely enhance visual realism. Existing methods can be broadly categorized according to how semantic supervision is obtained during generation, including Image-to-Mask, Mask-to-Image, and Image-Mask joint generation. In Image-to-Mask pipelines\citep{Re1_2023diffumask}\citep{Re1_Dataset_diffusion}\citep{Re1_SDS}, images are first generated or augmented, and semantic masks are subsequently obtained through segmentation models, pseudo-labeling strategies, or foundation-model-assisted annotation\citep{ren2024grounded}\citep{clip}. These methods are flexible and can benefit from powerful generative models, but their label quality is bound by the reliability of the subsequent mask prediction stage. In contrast, Mask-to-Image methods\citep{Re1_ye2024seggen}\citep{Re1_yang2023freemask} synthesize images conditioned on given semantic layouts or class masks. This paradigm provides explicit control over spatial structures and class compositions, but typically relies on existing masks and mainly diversifies visual appearance under fixed label structures, thereby limiting scene-level diversity. More recently, Image-Mask joint generation methods\citep{wang2026jodiffusion}\citep{Re1_joint_zhang2025paired} attempt to synthesize images and their corresponding semantic labels simultaneously through shared latent modeling or coupled optimization objectives. For example, JoDiffusion\citep{wang2026jodiffusion} introduces a text-prompt-driven joint generation framework for images and pixel-wise semantic labels, where a unified latent space, joint diffusion process, and mask refinement strategy are employed to synthesize diverse and highly consistent image-label pairs. Since image content and pixel-wise labels can be constrained within a unified generation process, joint generation is better suited to segmentation-oriented data augmentation. Nevertheless, most existing semantic dataset generation methods remain designed for image-label pairs, especially in the visible domain, with limited consideration of heterogeneous sensing modalities such as infrared imagery.

\subsection{Visible-to-Infrared Image Translation}

Visible-to-infrared image translation\citep{dai2025diffusion}\citep{mao2026pid}\citep{ran2025diffv2ir} aims to synthesize infrared images from visible images, thereby alleviating the scarcity of thermal data and expanding modality availability for multimodal perception. Early studies commonly formulate this task as image-to-image translation and learn cross-domain mappings through Variational Autoencoders (VAEs)\citep{vae_hwang2020variational}\citep{vae_kingma2013auto_vae} or Generative Adversarial
Networks (GANs)\citep{vae_hwang2020variational}\citep{GAN_isola2017image}\citep{GAN_wang2018high}\citep{zhu2017unpaired}. To better preserve scene structures during modality transfer, subsequent methods introduce additional constraints such as edge guidance\citep{lee2023edge_gan}, structural similarity\citep{ozkanouglu2022infragan}, representation
learning\citep{han2024dr}, or multi-scale feature alignment\citep{sun2023vq}. More recently, diffusion-based\citep{T2i-adapter}\citep{dai2025diffusion}\citep{chen2026any2any} and conditional generative methods\citep{yang2025s} have further improved infrared synthesis by leveraging stronger generative priors to model the target infrared distribution. These studies demonstrate the feasibility of synthesizing infrared-like images from visible observations and provide useful tools for modality expansion. However, most existing visible-to-infrared translation methods are essentially designed as bimodal mapping models. They usually focus on generating visually plausible infrared images, while their effectiveness and direct usability for downstream tasks are rarely explicitly optimized.

% To augment VIS-IR-Label training data for complex scenes, a straightforward solution is to cascade the above two lines of methods: generating a visible image and its semantic label, translating the visible image into the infrared domain, and assembling the three outputs into a triplet. However, such a cascaded pipeline decomposes the unified triplet generation problem into several loosely coupled stages. The generated label may be inconsistent with the translated infrared structures; the infrared image is synthesized from an intermediate visible output rather than constrained by triplet-level global semantics, and errors may accumulate across image generation, label acquisition, and modality translation. This issue becomes more pronounced under few-shot paired supervision, where independently trained modules cannot fully exploit limited VIS-IR-Label correspondences to learn global cross-modal constraints. In contrast, UniTriGen formulates the visible image, infrared image, and semantic label as a unified generation target, enabling spatial alignment, semantic consistency, and modality complementarity to be jointly modeled within a single diffusion process.

To generate VIS-IR-Label training triplets for complex scenes, a straightforward strategy is to combine image-label generation with visible-to-infrared translation. Such a cascaded pipeline first synthesizes visible images and their semantic labels and then translates the visible images into the infrared domain. However, this formulation decomposes triplet generation into loosely coupled local conditional processes, where the label and infrared modality are generated at different stages rather than being jointly constrained within a unified diffusion modeling process. As a result, the assembled triplets may suffer from spatial misalignment, semantic mismatch, boundary inconsistency, and cross-modal detail conflicts. This issue becomes more severe when learning from only a limited number of paired VIS-IR-Label triplets, since the scarce cross-modal correspondences make it difficult for independently trained modules to capture global cross-modal consistency. In contrast, UniTriGen treats VIS, IR, and Label as a unified generation target and models them within a single diffusion framework, enabling joint enforcement of spatial alignment, semantic consistency, and modality complementarity.
% \vspace{-6pt}

\section{Methodology}
\label{gen_inst}

\subsection{Problem Setup}
We investigate multi-source aligned triplet generation under a few-shot learning setting. Given a limited real paired dataset $\mathcal{D}_{real}=\{(V_i, I_i, L_i, T_i)\}_{i=1}^{N}$, where $V_i$, $I_i$, and $L_i$ represent the visible image, infrared image, and semantic label from the same scene, respectively, and $T_i$ denotes the corresponding text prompt. Our goal is to learn a text-conditioned triplet generation model $G_\theta: T\rightarrow (V, I, L)$, which can generate spatially aligned, semantically consistent, and modality complementary VIS-IR-Label triplets in an end-to-end manner solely from text prompts. 

The synthetic dataset is denoted as $\mathcal{D}_{syn}=\{(\hat{V}_j,\hat{I}_j,\hat{L}_j)\}_{j=1}^{N_{syn}}$. These samples are expected to satisfy two fundamental requirements. First, the triplets outputs should exhibit strict consistency in scene layout, object locations, and class semantics. Second, the generated labels should be trainable and directly serve as supervision signals for downstream RGB-T semantic segmentation models. Therefore, the final objective is to expand the training set using limited real paired data in combination with a synthetic dataset ($\mathcal{D}_{real}\cup\mathcal{D}_{syn}$), thereby enhancing the performance of downstream RGB-T semantic segmentation models.

\FloatBarrier  % 阻止之前的浮动体越过这里
\begin{figure}[htbp]
    \centering
    \includegraphics[width=\textwidth]{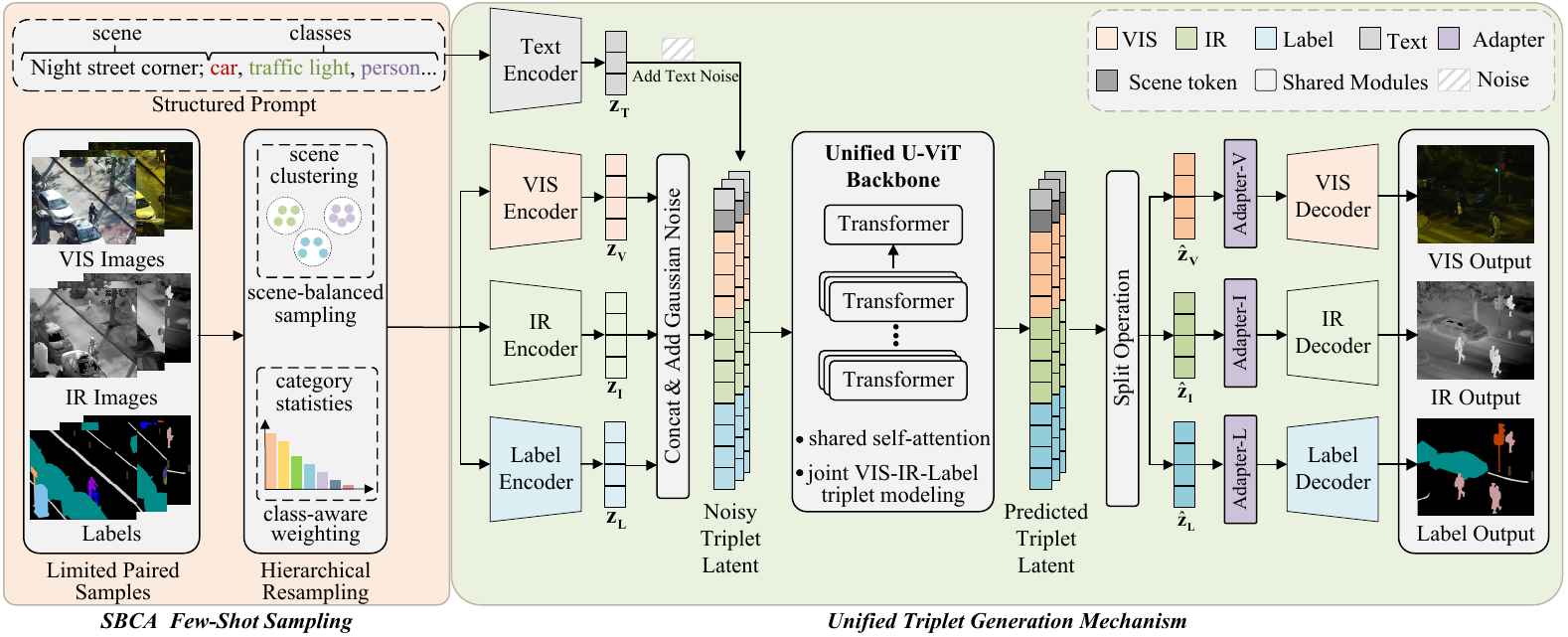}
    \caption{Overview of the UniTriGen framework. The framework consists of two main components: Unified Triplet Generation Mechanism and SBCA Few-Shot Sampling.}
    \label{framwork}
\end{figure}
% \FloatBarrier  % 阻止之前的浮动体越过这里

\subsection{Unified Triplet Generation Mechanism}
\label{sec:unified_triplet_generation}

To enable end-to-end text-to-VIS-IR-Label Aligned triplet generation, we construct a \textbf{Unified Triplet Generation Mechanism} based on the UniDiffuser\citep{unidiffuser} architecture, which integrates joint cross-modal diffusion modeling and modality-specific latent calibration into a single generation process. The core idea is to first encode the visible image, infrared image, and semantic label into a shared latent space and jointly model them within a unified diffusion process, so that the three modalities can share scene-level structural and semantic constraints throughout generation. Then, before decoding, lightweight modality-specific residual adapters calibrate the predicted latents to better match the decoding characteristics of each modality. In this way, unified triplet generation simultaneously preserves cross-modal consistency and improves modality-dependent fidelity.

% Unlike cascaded paradigms that generate one or two modalities first and then derive the remaining ones, our method directly learns the text-conditioned joint distribution of the complete VIS-IR-Label triplet. This unified formulation avoids decomposing the target distribution into several local conditional generation stages, thereby reducing layout shifts, boundary inconsistencies, semantic misalignment, and cross-modal detail conflicts.

\noindent\textbf{Joint triplet latent diffusion.}
Given a text prompt $T$, we employ the CLIP\citep{clip} text encoder $\mathcal{E}_{T}$ to obtain the text latent representation $z_T=\mathcal{E}_{T}(T)$. For an aligned triplet $(V, I, L)$, we use modality-specific VAE encoders $E_V$, $E_I$, and $E_L$ to encode the visible image, infrared image, and semantic label, respectively, and concatenate their latent variables into a unified triplet latent representation:
\begin{equation}
    z_V = E_V(V), \quad
    z_I = E_I(I), \quad
    z_L = E_L(L), \quad
    z_0 = [z_V;z_I;z_L].
\end{equation}
This representation enables the model to capture visible appearance, infrared thermal cues, and label boundaries within a unified generation space, rather than treating them as independent targets.

In the forward diffusion process, Gaussian noise is injected into $z_0$ to obtain the noisy latent variable at timestep $t$:
\begin{equation}
    q(z_t \mid z_0) = \mathcal{N}\left(\sqrt{\bar{\alpha}_t} z_0, (1-\bar{\alpha}_t)\mathbf{I}\right), \quad
    z_t = \sqrt{\bar{\alpha}_t} z_0 + \sqrt{1-\bar{\alpha}_t} \epsilon, \quad \epsilon \sim \mathcal{N}(0,\mathbf{I}).
\end{equation}
This joint noising scheme encourages the network to learn spatial correspondence and semantic coupling across modalities.

During reverse denoising, the text latent $z_T$ serves as the sole external condition and is fed together with the noisy triplet latent into a unified U-ViT backbone. Through self-attention over the unified token sequence, the network models interactions between text and the three modality-specific latent components and predicts the injected noise $\epsilon_{\theta}(z_t,z_T,t)$. The reverse transition is defined as
\begin{equation}
    p_{\theta}(z_{t-1}\mid z_t,z_T) = \mathcal{N}\left(\mu_{\theta}(z_t,z_T,t), \sigma_t^2\mathbf{I}\right),
\end{equation}
where the denoised mean is computed by
\begin{equation}
    \mu_{\theta}(z_t,z_T,t) = \frac{1}{\sqrt{\alpha_t}}\left(z_t - \frac{1-\alpha_t}{\sqrt{1-\bar{\alpha}_t}} \epsilon_{\theta}(z_t,z_T,t) \right).
\end{equation}

The baseline is optimized with the standard noise prediction objective:
\begin{equation}
    \mathcal{L}_{\mathrm{denoise}} = \mathbb{E}_{t,z_0,\epsilon}\left[\|\epsilon_{\theta}(z_t,z_T,t) - \epsilon\|_2^2\right].
\end{equation}

\noindent\textbf{Modality-specific latent calibration.}
Although the unified diffusion backbone jointly models VIS, IR, and Label in a shared latent space, the three modalities have distinct decoding requirements. VIS emphasizes color, texture, and structural details; IR highlights thermal responses and salient contours; and Label requires discrete semantic regions with clear object boundaries. Therefore, directly feeding the backbone-predicted latents into their corresponding decoders may cause residual discrepancies between the predicted latents and the target modality-specific decoding manifolds.

To address this issue while preserving the unified generation process, we introduce lightweight modality-specific residual adapters after joint denoising and before triplet decoding. Inspired by Any2Any~\citep{chen2026any2any}, each adapter performs a residual calibration on the corresponding predicted latent. 
We denote the predicted latent of modality $m$ as $\hat{z}_m$, where $m \in \{V,I,L\}$. The calibrated latent is defined as:
\begin{equation}
    \tilde{z}_m
    =
    \hat{z}_m
    +
    A_{\phi}^{m}(\hat{z}_m),
    \quad
    m \in \{V,I,L\},
\end{equation}
where $A_{\phi}^{m}$ denotes the lightweight adapter for modality $m$, and $\phi$ denotes its learnable parameters. The calibrated latents are then decoded by their corresponding modality decoders:
\begin{equation}
    \hat{x}_m = D_m(\tilde{z}_m),
    \quad
    m \in \{V,I,L\},
\end{equation}
where $\hat{x}_V$, $\hat{x}_I$, and $\hat{x}_L$ correspond to the generated visible image, infrared image, and semantic label, respectively. $D_m$ denotes the decoder corresponding to modality $m$.

During training, the adapters are supervised in the latent space by encouraging the calibrated latent to approximate the corresponding ground-truth modality latent. To prevent the adapter loss from disturbing the unified diffusion backbone, stop-gradient is applied to the backbone-predicted latent when optimizing the adapters:
\begin{equation}
    \mathcal{L}_{\mathrm{calib}}
    =
    \sum_{m \in \{V,I,L\}}
    \left\|
    \mathrm{sg}(\hat{z}_m)
    +
    A_{\phi}^{m}(\mathrm{sg}(\hat{z}_m))
    -
    z_m
    \right\|_2^2,
\end{equation}
where $z_m = E_m(x_m)$ denotes the ground-truth latent of modality $m$, and $\mathrm{sg}(\cdot)$ denotes the stop-gradient operation. The overall objective of unified triplet generation is:
\begin{equation}
    \mathcal{L}_{\mathrm{UTG}}
    =
    \mathcal{L}_{\mathrm{denoise}}
    +
    \lambda
    \mathcal{L}_{\mathrm{calib}},
\end{equation}
where $\lambda$ balances joint triplet diffusion modeling and modality-specific latent calibration.

During inference, VIS, IR, and Label are first jointly generated through the unified reverse diffusion process. The denoised triplet latent is then calibrated once by the modality-specific adapters and decoded into the final triplet outputs. Since the adapters are only applied after denoising and do not participate in iterative diffusion sampling, they introduce negligible computational overhead. 
Overall, the unified triplet generation mechanism enables UniTriGen to produce spatially aligned, semantically consistent, and modality-complementary VIS-IR-Label triplets from text prompts within a single coherent generation process.

\subsection{SBCA Few-Shot Sampling}
\label{sec:scene_balanced_sampling}

When only a limited number of paired triplets are available for training, the training data often exhibit highly imbalanced scene and semantic class distributions. Such imbalance biases the generative model toward frequent scenes and common classes, leading to inadequate representation of low-frequency scenes and rare classes. Consequently, the diversity of the synthetic triplets is substantially limited. To mitigate this issue, we propose a \textbf{S}cene-\textbf{B}alanced and \textbf{C}lass-\textbf{A}ware (\textbf{SBCA}) \textbf{Few-Shot Sampling} strategy, which induces more balanced sampling distributions over scenes and classes through structured text condition, explicit scene encoding, and hierarchical sample resampling.

\noindent\textbf{Structured scene-class prompt.}
We structure the text prompt of each training sample into a scene-class format. Specifically, for each sample $i$, its text representation $T_i$ is constructed as $[d_i^s; d_i^c]$, where $d_i^s$ denotes the scene description and $d_i^c$ denotes the set of semantic classes appearing in the sample. For example, a sample can be written as:
\textit{Nighttime parking area adjacent to a dense forest background; car, traffic light, pole, curve, person.}
This format explicitly injects both global scene context and local object classes into the text condition, enabling the model to perceive scene background and object semantics simultaneously.

\noindent\textbf{Explicit scene encoding.}
To disentangle scene semantics from class semantics in the text embedding space, we introduce explicit scene encoding. We first employ the Places365-CNNs~\citep{zhou2017places365} classification model to extract scene representations, and cluster the training set into $K$ scene groups:
\begin{equation}
    \mathcal{D}_{real}
    =
    \{
    \mathcal{G}_1,
    \mathcal{G}_2,
    \ldots,
    \mathcal{G}_K
    \}.
\end{equation}
For sample $i$, its scene identity is denoted as $s_i \in \{1,\ldots,K\}$. During text encoding, we append a learnable scene token to the original text representation, giving the augmented text latent:
\begin{equation}
    \tilde{z}_{T_i}
    =
    [\mathcal{E}_T(T_i); e_{s_i}],
\end{equation}
where $e_{s_i}$ is the learnable embedding of scene $s_i$. In practice, $\tilde{z}_{T_i}$ replaces $z_T$ as the text condition in unified triplet generation, allowing the diffusion backbone to receive both class-aware semantic prompts and explicit scene-level guidance.

\noindent\textbf{Scene-balanced and class-aware resampling.}
During training, we implement a hierarchical resampling mechanism: first, balancing the overall sampling probability across scenes, and then weighting samples within each scene according to class rarity. For each class $c$, let $N_c$ denote the number of training samples containing this class. If sample $i$ belongs to scene group $\mathcal{G}_k$, its final sampling weight is defined as:
\begin{equation}
    W_i
    =
    \frac{1}{K}
    \cdot
    \frac{r_i}{\sum_{j\in\mathcal{G}_k} r_j},
    \quad
    \text{where}
    \quad
    r_i
    =
    \epsilon
    +
    \max_{c\in\mathcal{C}_i} w_c,
    \quad
    w_c
    =
    \frac{1}{N_c^\alpha}.
\end{equation}
Here, $K$ denotes the number of scene groups, $\mathcal{G}_k$ denotes the scene group to which sample $i$ belongs, $\mathcal{C}_i$ denotes the class set of sample $i$, $\alpha$ is a smoothing coefficient, and $\epsilon$ ensures a non-zero probability for every sample. This mechanism enhances supervision for tail scenes and classes, yielding a more balanced training distribution for VIS-IR-Label triplet generation.

\section{Experiments and Results}
\subsection{Datasets and Experimental Settings}
\label{Datasets and Experimental Settings}
\textbf{Datasets.}
We evaluate UniTriGen on two public RGB-T semantic segmentation datasets: \textbf{SemanticRT}\citep{semanticrt} and \textbf{PST900}\citep{pst900}. 
% SemanticRT contains 11,371 aligned RGB-thermal image pairs with pixel-level annotations for 13 semantic classes, including the background class. It consists of 6,830 training samples, 1,705 validation samples, and 2,836 testing samples, based on the official split. PST900 is a dataset designed for robotic perception in underground environments, with 597 training pairs and 288 testing pairs, covering five classes: fire extinguisher, backpack, hand drill, survivor, and background.
To simulate practical scenarios with limited paired triplet data, we conduct experiments under a few-shot setting. Specifically, for \textbf{SemanticRT}, we sample \textbf{5\%} of the dataset, resulting in 342 training samples, 85 validation samples, and 142 testing samples. For \textbf{PST900}, we use \textbf{50\%} of the dataset, with 299 training samples and 144 testing samples. All methods are evaluated with the same sampled splits for a fair comparison.

% We evaluate UniTriGen on two public RGB-T semantic segmentation datasets, i.e., SemanticRT and PST900. \textbf{SemanticRT} contains 11,371 aligned RGB-thermal image pairs with pixel-level annotations for 13 semantic classes, including the background class. Following the official split, it consists of 6,830 training samples, 1,705 validation samples, and 2,836 testing samples. \textbf{PST900} is an RGB-thermal semantic segmentation dataset designed for robotic perception in challenging underground environments. Its latest version provides 597 training pairs and 288 testing pairs with manually annotated pixel-wise semantic labels. The dataset covers five classes, including four foreground classes, i.e., fire extinguisher, backpack, hand drill, and survivor, as well as the background class.

% To better reflect practical scenarios where only limited paired RGB-thermal data are available, we conduct experiments under a few-shot paired-data setting. Specifically, for \textbf{SemanticRT}, we randomly sample \textbf{5\% of the original dataset}, resulting in 342 training samples, 85 validation samples, and 142 testing samples. For \textbf{PST900}, we use \textbf{50\% of the dataset}, including 299 training samples and 144 testing samples. All compared methods adopt the same sampled splits to ensure a fair comparison.

\textbf{Experimental Settings.}
For both datasets, visible images, thermal images, and semantic labels are resized to $512 \times 512$ for training. We use the AdamW\citep{AdamW} optimizer for all training stages of the VAE and diffusion models. Random horizontal flipping is adopted as the basic data augmentation strategy. All compared methods follow the same data preprocessing pipeline and training settings. 
% More details on the architectures and hyperparameters of the VAE and diffusion models are provided in the supplementary material\ref{Implementation Details}.

\subsection{Main Comparison with Existing Pipelines}

% \subsection{Comparison with Cascaded Generation Paradigms}

To the best of our knowledge, text-driven generation of aligned VIS-IR-Label triplets from limited paired training data has not been explicitly studied before. Therefore, we construct three representative cascaded alternatives for comparison, namely Seq-Cascade, JointLabel-Cascade, and TriCond-Cascade. To ensure fairness, the text-to-image generation module and the joint generation module in the three cascaded schemes are both implemented based on the UniDiffuser\citep{unidiffuser} architecture, while the Visible-to-Infrared image translation module is implemented based on DiffV2IR\citep{ran2025diffv2ir}.

\paragraph{Quantitative Results}
% \subsubsection{Quantitative Results}
Table~\ref{tab:main_compare_on_different_segmentation_methods} compares our method with three representative cascaded pipelines on the SemanticRT and PST900 datasets. On the 5\% SemanticRT setting, where only 342 real training pairs are available, adding an equal number of synthetic triplets generated by UniTriGen consistently improves all downstream RGB-T segmentation models. Similar trends can be observed on the 50\% PST900 setting, where 299 synthetic triplets are added to 299 real training pairs. UniTriGen improves the raw-data baseline from 78.51 to 83.96, 74.96 to 79.31, 85.86 to 87.06, and 86.42 to 87.29 across the four downstream architectures, respectively. These results demonstrate that UniTriGen-generated triplets provide effective, model-agnostic data augmentation with limited paired supervision.
Compared with the cascaded alternatives, UniTriGen achieves consistently superior performance in both the synthetic-only and real-plus-synthetic settings. When combined with real data, UniTriGen consistently achieves the highest mIoU among all compared methods across different downstream models. On PST900, the advantage is more pronounced. These consistent improvements indicate that cascaded pipelines, although able to assemble VIS-IR-Label triplets, suffer from accumulated errors and insufficient joint cross-modal constraints. In contrast, UniTriGen directly models the joint distribution of VIS, IR, and Label within a unified diffusion process, leading to better spatial alignment, semantic consistency, and downstream training utility. 
% Additional quantitative results can be found in the supplementary material \ref{Addition Comparision}.

\begin{table*}[t]
\centering
\setlength{\tabcolsep}{4.6pt}
\renewcommand{\arraystretch}{1.18}
\resizebox{\textwidth}{!}{
\begin{tabular}{l l l c c c c c c}
\toprule

\multirow{2}{2.2cm}{\centering\arraybackslash\textbf{Downstream Methods}}
& \multirow{2}{*}{\textbf{Backbone}}
& \multirow{2}{*}{\textbf{Augmentation Methods}}
& \multicolumn{3}{c}{\textbf{5\% SemanticRT Dataset}}
& \multicolumn{3}{c}{\textbf{50\% PST900 Dataset}} \\

\cmidrule(lr){4-6} \cmidrule(lr){7-9}

&
&
& \textbf{Data Size (Pairs)}
& \textbf{mIoU (Syn only)}
& \textbf{mIoU (Real+Syn)}
& \textbf{Data Size (Pairs)}
& \textbf{mIoU (Syn only)}
& \textbf{mIoU (Real+Syn)} \\

\midrule

\multirow{5}{*}{\makecell[l]{SemanticRT\citep{semanticrt}}}
& \multirow{5}{*}{ResNet152}
& Raw Data
& 342
& \multicolumn{2}{c}{66.62}
& 299
& \multicolumn{2}{c}{78.51} \\
\cmidrule(lr){3-9}

&
& Seq-Cascade
& 684 & 41.81 & 71.59
& 598 & 39.87 & 79.76 \\

&
& JointLabel-Cascade
& 684 & 43.22 & 71.78
& 598 & 35.77 & 78.10 \\

&
& TriCond-Cascade
& 684 & \underline{50.76} & \underline{73.02}
& 598 & \underline{40.83} & \underline{81.95} \\

&
& \cellcolor{oursgray}UniTriGen (Ours)
& \cellcolor{oursgray}684
& \cellcolor{oursgray}\textbf{54.16}
& \cellcolor{oursgray}\textbf{73.12}
& \cellcolor{oursgray}598
& \cellcolor{oursgray}\textbf{41.39}
& \cellcolor{oursgray}\textbf{83.96} \\

\midrule

\multirow{5}{*}{\makecell[l]{M-SpecGene\citep{M-SpecGene}}}
& \multirow{5}{*}{UperNet}
& Raw Data
& 342
& \multicolumn{2}{c}{69.38}
& 299
& \multicolumn{2}{c}{74.96} \\
\cmidrule(lr){3-9}

&
& Seq-Cascade
& 684 & 43.19 & 70.71
& 598 & 49.41 & 75.50 \\

&
& JointLabel-Cascade
& 684 & 43.63 & 70.99
& 598 & 49.17 & 76.00 \\

&
& TriCond-Cascade
& 684 & \underline{51.79} & \underline{71.17}
& 598 & \underline{49.46} & \underline{76.76} \\

&
& \cellcolor{oursgray}UniTriGen (Ours)
& \cellcolor{oursgray}684
& \cellcolor{oursgray}\textbf{58.90}
& \cellcolor{oursgray}\textbf{71.57}
& \cellcolor{oursgray}598
& \cellcolor{oursgray}\textbf{65.48}
& \cellcolor{oursgray}\textbf{79.31} \\

\midrule

\multirow{5}{*}{\makecell[l]{Sigma\citep{sigma}}}
& \multirow{5}{*}{Mamba}
& Raw Data
& 342
& \multicolumn{2}{c}{73.46}
& 299
& \multicolumn{2}{c}{85.86} \\
\cmidrule(lr){3-9}

&
& Seq-Cascade
& 684 & 42.72 & 73.04
& 598 & 51.76 & 85.99 \\

&
& JointLabel-Cascade
& 684 & 47.56 & 73.45
& 598 & 46.37 & 85.25 \\

&
& TriCond-Cascade
& 684 & \underline{53.20} & \underline{73.65}
& 598 & \underline{52.88} & \underline{86.48} \\

&
& \cellcolor{oursgray}UniTriGen (Ours)
& \cellcolor{oursgray}684
& \cellcolor{oursgray}\textbf{58.11}
& \cellcolor{oursgray}\textbf{74.36}
& \cellcolor{oursgray}598
& \cellcolor{oursgray}\textbf{53.61}
& \cellcolor{oursgray}\textbf{87.06} \\

\midrule

\multirow{5}{*}{\makecell[l]{MilNet\citep{MiLNet}}}
& \multirow{5}{*}{Segformer}
& Raw Data
& 342
& \multicolumn{2}{c}{69.73}
& 299
& \multicolumn{2}{c}{86.42} \\
\cmidrule(lr){3-9}

&
& Seq-Cascade
& 684 & 40.90 & 69.92
& 598 & 54.50 & 85.34 \\

&
& JointLabel-Cascade
& 684 & 48.70 & 70.70
& 598 & 47.80 & 84.30  \\

&
& TriCond-Cascade
& 684 & \underline{55.60} & \underline{71.02}
& 598 & \underline{55.60} & \underline{85.37} \\

&
& \cellcolor{oursgray}UniTriGen (Ours)
& \cellcolor{oursgray}684
& \cellcolor{oursgray}\textbf{57.30}
& \cellcolor{oursgray}\textbf{71.45}
& \cellcolor{oursgray}598
& \cellcolor{oursgray}\textbf{64.90}
& \cellcolor{oursgray}\textbf{87.29} \\

\bottomrule
\end{tabular}
}
\caption{Quantitative comparison of different methods on SemanticRT and PST900 datasets.
The best results are highlighted in \textbf{bold}, while the second-best outcomes are denoted by \underline{\textit{underlined italic}}.}
% \vspace{-6pt}
\label{tab:main_compare_on_different_segmentation_methods}
\end{table*}
% \vspace{-6pt}

\begin{figure}[htbp]
    \centering
    \includegraphics[width=\textwidth]{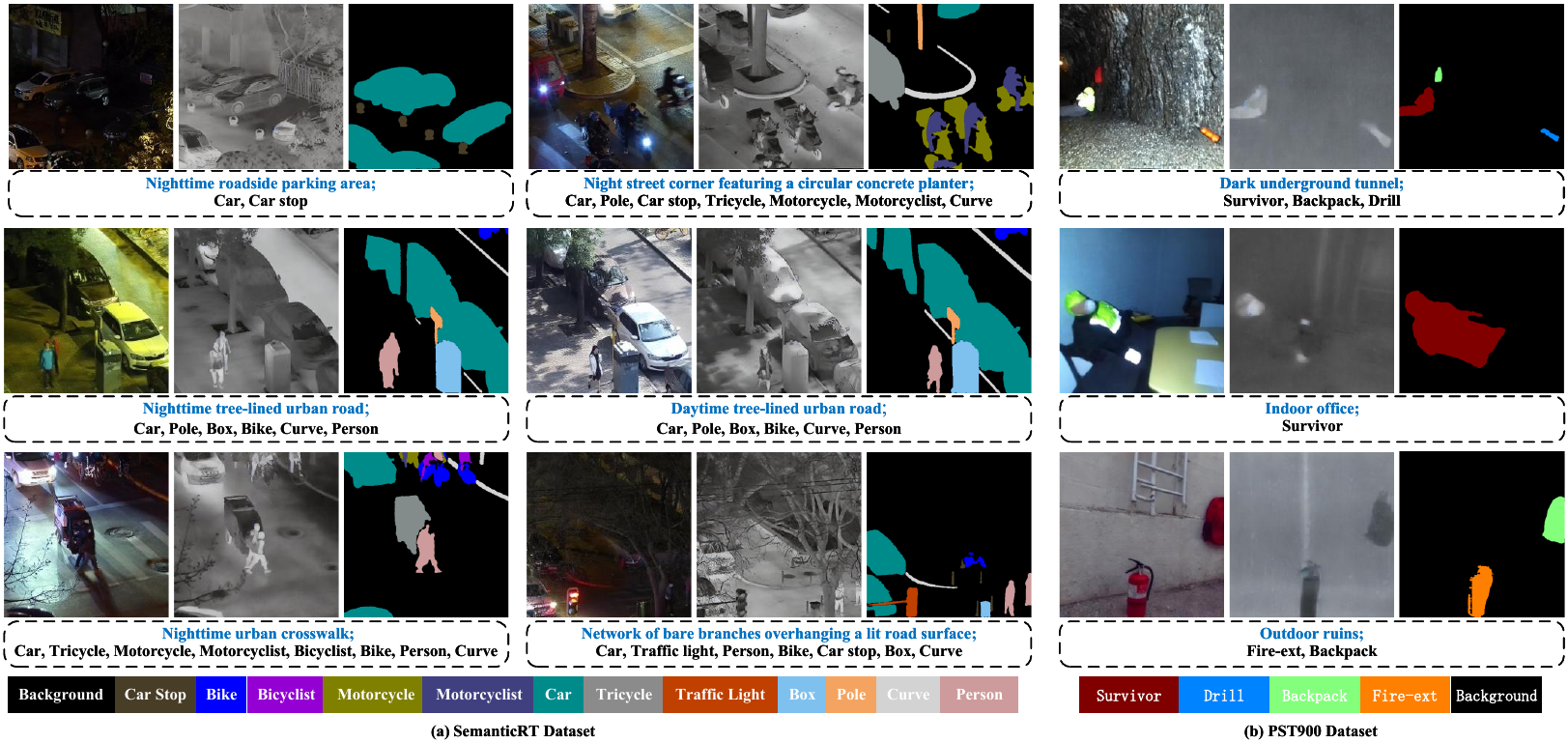}
    \caption{Visualization of UniTriGen generation results on the SemanticRT dataset and PST900 dataset. The text prompt consists of \textcolor[rgb]{0.00,0.44,0.75}{scene} and \textcolor[rgb]{0.00,0.00,0.00}{cotegories}.}
    \label{UniTriGen_viaualization}
\end{figure}

\paragraph{Qualitative Results}
% \subsubsection{Qualitative Results}
\label{subsubsec:Quantitative Results}
Figure~\ref{UniTriGen_viaualization} visualizes the VIS-IR-Label triplets generated by UniTriGen on SemanticRT and PST900. Across diverse scenes and classes, UniTriGen produces aligned triplets with clear spatial alignment and semantic consistency among the three modalities. The generated VIS images preserve realistic scene appearance and structural layouts, while the IR images exhibit modality-specific thermal characteristics, such as salient human targets and weakened texture responses, rather than simply copying visible-domain patterns. Meanwhile, the generated labels are well aligned with object boundaries and remain accurate for small or thin structures, such as poles, car stop, drills, and fire extinguishers. Notably, UniTriGen can distinguish visually or semantically similar classes, including Tricycle, Motorcycle, and Bike, as well as Motorcyclist, Bicyclist, and Person. These results indicate that the proposed unified generation framework can synthesize high-quality VIS-IR-Label triplets with strong cross-modal coherence.

Figure~\ref{Comparision_visualization} provides a qualitative comparison between UniTriGen and the cascaded variants. Overall, the cascaded pipelines show evident cross-modal inconsistency due to their multi-stage generation process. In JointLabel-Cascade and TriCond-Cascade, once a class is missed in the generated label, such as the person in the dark region, the corresponding object is also absent or weakened in the synthesized IR image, indicating that errors are propagated across stages. Moreover, JointLabel-Cascade sometimes produces IR results that resemble appearance-level tone transfer rather than physically plausible thermal imaging; for example, the covered parked car exhibits unrealistic thermal responses. Seq-Cascade further suffers from noisy semantic predictions, since its final labels heavily depend on the quality of the previously generated VIS and IR images. In contrast, UniTriGen directly models VIS, IR, and Label as a unified triplet, producing more consistent object presence, more realistic IR modality characteristics, and better-aligned semantic labels. 

\FloatBarrier  % 阻止之前的浮动体越过这里
\vspace{-8pt}

\subsection{Ablation Studies}
\begin{figure*}[t]
\centering

% ===================== Left: Image =====================
\begin{minipage}[t]{0.54\textwidth}

\vspace{0pt}
\centering
\includegraphics[width=\linewidth,height=0.25\textheight,keepaspectratio]{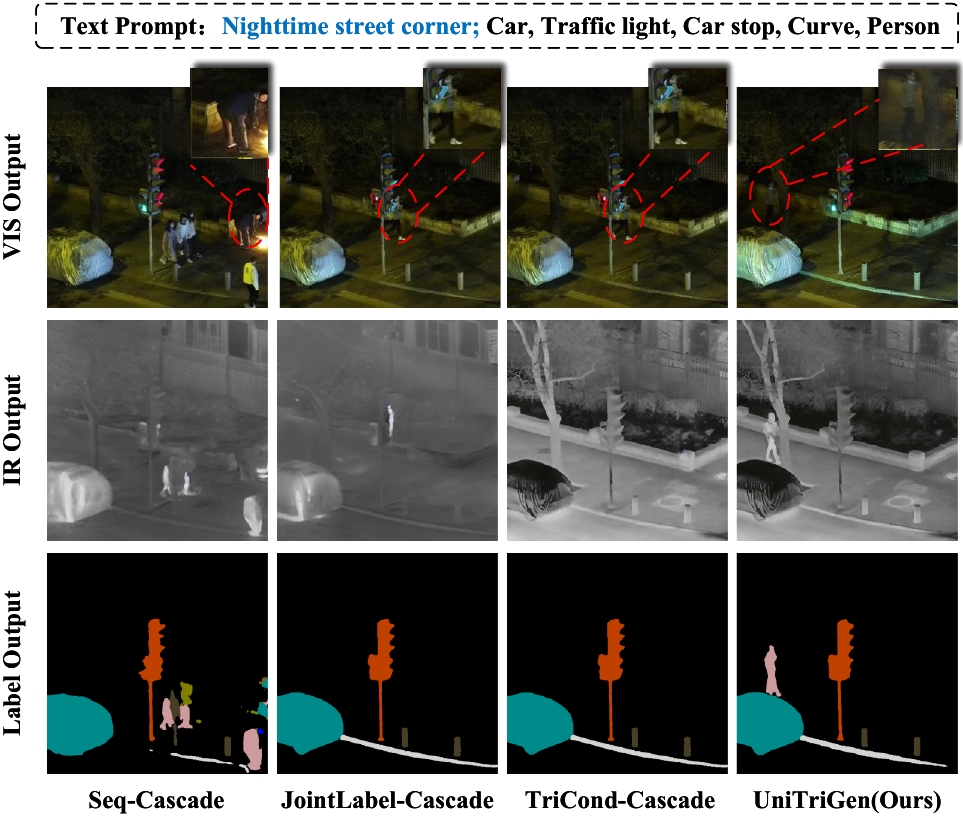}
\captionsetup{font=small}
\captionof{figure}{Qualitative comparison of different VIS-IR-Label generation pipelines on the SemanticRT dataset.}
\label{Comparision_visualization}
\end{minipage}
\hfill
% ===================== Right: Two Tables =====================
\begin{minipage}[t]{0.44\textwidth}
\vspace{0pt}
\centering

\setlength{\tabcolsep}{4.5pt}
\renewcommand{\arraystretch}{1.10}
\resizebox{\linewidth}{!}{
\begin{tabular}{cc cc}
\toprule
\multicolumn{2}{c}{\textbf{Method Components}} 
& \multicolumn{2}{c}{\textbf{Downstream Performance}} \\
\cmidrule(lr){1-2} \cmidrule(lr){3-4}
\makecell[c]{\textbf{UniTriGen}\\\textbf{Mechanism}} 
& \makecell[c]{\textbf{SBCA Few-Shot}\\\textbf{Sampling}} 
& \textbf{5\% SemanticRT} 
& \textbf{50\% PST900} \\
\midrule
$\checkmark$ &  
& 70.79 & 78.79 \\

\rowcolor{rowgray}
$\checkmark$ & $\checkmark$ 
& 71.57 & 79.31  \\
\bottomrule
\end{tabular}
}
\captionof{table}{Ablation results for different components of the proposed method, where “UniTriGen Mechanism” denotes the unified triplet generation mechanism.}
\label{tab:ablation_main}

\vspace{1.0em}

% --------------------- Table 3 ---------------------
\setlength{\tabcolsep}{4.2pt}
\renewcommand{\arraystretch}{1.10}
\resizebox{\linewidth}{!}{
\begin{tabular}{l
                S[table-format=2.2]
                S[table-format=2.2]
                l
                S[table-format=2.2]
                S[table-format=2.2]}
\toprule
\multicolumn{3}{c}{SemanticRT} & 
\multicolumn{3}{c}{PST900} \\
\cmidrule(lr){1-3} \cmidrule(lr){4-6}
\ {Ratio} & {Real} & {Real+Syn} &
\ {Ratio} & {Real} & {Real+Syn} \\
\midrule
5\%  & 69.38 & 71.57 & 50\%  & 74.96 & 79.31 \\
10\% & 71.65 & {73.62} & 75\%  & 76.52 & {80.19} \\
25\% & 77.35 & {78.28} & 100\% & 79.71 & {81.92} \\
\bottomrule
\end{tabular}
}
\captionof{table}{Performance comparison under different real-data ratios.}
\label{tab:ratio_comparison}

\end{minipage}

\end{figure*}

Table~\ref{tab:ablation_main} presents the ablation study results of the proposed components in UniTriGen. 
Without incorporating synthetic data, the M-SpecGene~\citep{M-SpecGene} method achieves 69.38 on the original 5\% SemanticRT setting and 74.96 on the original 50\% PST900 setting. 
When applying the unified triplet generation mechanism, the performance increases to 70.79 and 78.79 on SemanticRT and PST900, respectively, demonstrating that jointly modeling VIS, IR, and label triplets provides an effective foundation for cross-modal data generation. 
Building upon this mechanism, introducing SBCA Few-Shot Sampling further improves the performance to 71.57 on SemanticRT and 79.31 on PST900, yielding gains of 0.78 and 0.52 points, respectively. 
These results indicate that scene-balanced and class-aware sampling can better exploit limited paired data and alleviate the class-imbalance issue under few-shot settings. 
Overall, the ablation results validate the effectiveness and complementarity of the proposed generation and sampling strategies.

\subsection{Extend Analysis}

\paragraph{Performance under Different Real-data Ratios}
\label{subsubsec:real_data_ratios}
While our main experiments use 5\% of SemanticRT and 50\% of PST900, UniTriGen is not tied to a fixed real-data ratio. To evaluate its robustness across data scales, we train UniTriGen with different ratios of paired real triplets and add the generated triplets to the corresponding real training sets for downstream evaluation. As shown in Table~\ref{tab:ratio_comparison}, UniTriGen consistently improves M-SpecGene~\citep{M-SpecGene} across all tested ratios, demonstrating its effectiveness under varying amounts of paired supervision. Notably, even with only 342 real triplets from 5\% SemanticRT and 299 real triplets from 50\% PST900, UniTriGen still produces synthetic triplets that improve downstream performance. This indicates that UniTriGen does not rely on large-scale paired supervision, which is crucial for RGB-T scenarios where aligned VIS-IR-Label triplets are scarce.

\begin{table}[t]
\centering
\scriptsize
\setlength{\tabcolsep}{3.0pt}
\renewcommand{\arraystretch}{0.92}
\setlength{\abovecaptionskip}{2pt}
\setlength{\belowcaptionskip}{2pt}

\begin{tabular}{c c c c c}
\toprule
\multirow{2}{*}{Datasets}
& \multicolumn{4}{c}{Scaling with Synthetic Data Size} \\
\cmidrule(lr){2-5}
& 1Real
& 1Real:1Syn
& 1Real:10Syn
& 1Real:20Syn \\
\midrule
5\% SemanticRT
& 69.38
& 71.57{\color{gaingreen}(+2.19$\uparrow$)}
& 72.00{\color{gaingreen}(+2.62$\uparrow$)}
& 72.13{\color{gaingreen}(+2.75$\uparrow$)} \\
50\% PST900
& 74.96
& 79.31{\color{gaingreen}(+4.35$\uparrow$)}
& 80.38{\color{gaingreen}(+5.42$\uparrow$)}
& 81.54{\color{gaingreen}(+6.58$\uparrow$)} \\
\bottomrule
\end{tabular}

\caption{Comparison of downstream performance under different real-to-synthetic data ratios.}
\label{tab:scaling}
% \vspace{-4pt}
\end{table}

\paragraph{Scaling with Synthetic Data Size}
\label{subsubsec:synthetic_scaling}
We further examine the impact of synthetic data scale on downstream performance. As shown in Table~\ref{tab:scaling}, under the 5\% SemanticRT setting, increasing the real-to-synthetic data ratio from 1:1 to 1:10 and 1:20 steadily improves the mIoU of M-SpecGene\citep{M-SpecGene} from 71.57 to 72.00 and 72.13, corresponding to gains of 2.19, 2.62, and 2.75 relative to the real-only baseline, respectively. Similar trends can be observed on the 50\% PST900 setting. These trends indicate that the aligned triplets generated by UniTriGen can consistently benefit downstream training as more synthetic data are incorporated. Compared with relying solely on limited real samples, larger synthetic datasets provide richer scene layouts, object appearances, and cross-modal correspondences, thereby leading to improved few-shot RGB-T semantic segmentation performance.

% \section{Conclusion}
% In this paper, we propose UniTriGen, a text-driven VIS-IR-Label triplet generation framework under limited paired triplet supervision. UniTriGen directly models VIS, IR, and Label within a unified generation process, addressing the difficulty of cascaded generation paradigms in maintaining cross-modal consistency. Its Unified Triplet Generation Mechanism jointly encodes the three modalities into a shared latent space, models the concatenated latent representation with a diffusion process, and integrates lightweight modality-specific residual adapters to preserve both cross-modal consistency and modality-specific fidelity. To mitigate generation bias caused by imbalanced scene and class distributions, we further design an SBCA few-shot sampling strategy, which induces a more balanced sampling distribution and improves the diversity of generated triplets. Extensive experiments on SemanticRT and PST900 demonstrate the effectiveness of UniTriGen.

\section{Conclusion}

In this paper, we propose UniTriGen, a text-driven VIS-IR-Label triplet generation framework for scenarios with only a limited number of paired VIS-IR-Label triplets available for training. UniTriGen directly models visible images, infrared images, and semantic labels within a unified generation process, thereby alleviating the difficulty of cascaded generation paradigms in maintaining cross-modal consistency. Specifically, the proposed Unified Triplet Generation Mechanism jointly encodes the three modalities into a shared latent space, models the concatenated latent representation through a diffusion process, and incorporates lightweight modality-specific residual adapters to preserve both cross-modal consistency and modality-specific fidelity. In addition, to mitigate generation bias caused by imbalanced scene and class distributions, we design an SBCA few-shot sampling strategy, which encourages a more balanced sampling distribution and improves the diversity of generated triplets. Extensive experiments on SemanticRT and PST900 demonstrate the effectiveness of UniTriGen in generating training-ready VIS-IR-Label triplets and improving downstream performance.

% Despite its promising performance, UniTriGen still has several limitations. First, the current framework does not explicitly model sensor-specific physical imaging processes. As a result, although the generated infrared images generally exhibit reasonable infrared characteristics, some thermal responses may deviate from real-world imaging properties, especially in scenes involving complex heat transfer, reflective materials, or ambiguous thermal boundaries. Second, the generated triplets are directly incorporated into downstream training without an explicit data refinement or quality filtering strategy. Consequently, a small portion of low-quality synthetic samples may introduce noise and partially limit the downstream effectiveness of the generated data. Future work will explore physics-aware infrared generation and automatic synthetic data selection or correction mechanisms to further improve the fidelity, reliability, and utility of VIS-IR-Label triplet generation.

{\small
\newpage
\bibliographystyle{unsrtnat}
\bibliography{references}

@inproceedings{Re1_2023diffumask,
  title={Diffumask: Synthesizing images with pixel-level annotations for semantic segmentation using diffusion models},
  author={Wu, Weijia and Zhao, Yuzhong and Shou, Mike Zheng and Zhou, Hong and Shen, Chunhua},
  booktitle={Proceedings of the IEEE/CVF International Conference on Computer Vision},
  pages={1206--1217},
  year={2023}
}

@inproceedings{Re1_SDS,
  title={A training-free synthetic data selection method for semantic segmentation},
  author={Tang, Hao and Yu, Siyue and Pang, Jian and Zhang, Bingfeng},
  booktitle={Proceedings of the AAAI Conference on Artificial Intelligence},
  volume={39},
  number={7},
  pages={7229--7237},
  year={2025}
}

@inproceedings{SD,
  title={High-resolution image synthesis with latent diffusion models},
  author={Rombach, Robin and Blattmann, Andreas and Lorenz, Dominik and Esser, Patrick and Ommer, Bj{\"o}rn},
  booktitle={Proceedings of the IEEE/CVF conference on computer vision and pattern recognition},
  pages={10684--10695},
  year={2022}
}

@inproceedings{controlnet,
  title={Adding conditional control to text-to-image diffusion models},
  author={Zhang, Lvmin and Rao, Anyi and Agrawala, Maneesh},
  booktitle={Proceedings of the IEEE/CVF international conference on computer vision},
  pages={3836--3847},
  year={2023}
}

@inproceedings{ruiz2023dreambooth,
  title={Dreambooth: Fine tuning text-to-image diffusion models for subject-driven generation},
  author={Ruiz, Nataniel and Li, Yuanzhen and Jampani, Varun and Pritch, Yael and Rubinstein, Michael and Aberman, Kfir},
  booktitle={Proceedings of the IEEE/CVF conference on computer vision and pattern recognition},
  pages={22500--22510},
  year={2023}
}

@article{labs2025flux,
  title={FLUX. 1 Kontext: Flow Matching for In-Context Image Generation and Editing in Latent Space},
  author={Labs, Black Forest and Batifol, Stephen and Blattmann, Andreas and Boesel, Frederic and Consul, Saksham and Diagne, Cyril and Dockhorn, Tim and English, Jack and English, Zion and Esser, Patrick and others},
  journal={arXiv preprint arXiv:2506.15742},
  year={2025}
}

@inproceedings{podellsdxl,
  title={SDXL: Improving Latent Diffusion Models for High-Resolution Image Synthesis},
  author={Podell, Dustin and English, Zion and Lacey, Kyle and Blattmann, Andreas and Dockhorn, Tim and M{\"u}ller, Jonas and Penna, Joe and Rombach, Robin},
  booktitle={The Twelfth International Conference on Learning Representations}
}

@article{wu2023datasetdm,
  title={Datasetdm: Synthesizing data with perception annotations using diffusion models},
  author={Wu, Weijia and Zhao, Yuzhong and Chen, Hao and Gu, Yuchao and Zhao, Rui and He, Yefei and Zhou, Hong and Shou, Mike Zheng and Shen, Chunhua},
  journal={Advances in Neural Information Processing Systems},
  volume={36},
  pages={54683--54695},
  year={2023}
}

@inproceedings{TextSSR,
  title={TextSSR: diffusion-based data synthesis for scene text recognition},
  author={Ye, Xingsong and Du, Yongkun and Tao, Yunbo and Chen, Zhineng},
  booktitle={Proceedings of the IEEE/CVF International Conference on Computer Vision},
  pages={17464--17473},
  year={2025}
}

@article{Text2Earth,
  title={Text2Earth: Unlocking text-driven remote sensing image generation with a global-scale dataset and a foundation model},
  author={Liu, Chenyang and Chen, Keyan and Zhao, Rui and Zou, Zhengxia and Shi, Zhenwei},
  journal={IEEE Geoscience and Remote Sensing Magazine},
  year={2025},
  publisher={IEEE}
}

@inproceedings{unidiffuser,
  title={One transformer fits all distributions in multi-modal diffusion at scale},
  author={Bao, Fan and Nie, Shen and Xue, Kaiwen and Li, Chongxuan and Pu, Shi and Wang, Yaole and Yue, Gang and Cao, Yue and Su, Hang and Zhu, Jun},
  booktitle={International Conference on Machine Learning},
  pages={1692--1717},
  year={2023},
  organization={PMLR}
}

@article{ran2025diffv2ir,
  title={Diffv2ir: visible-to-infrared diffusion model via vision-language understanding},
  author={Ran, Lingyan and Wang, Lidong and Wang, Guangcong and Wang, Peng and Zhang, Yanning},
  journal={arXiv preprint arXiv:2503.19012},
  year={2025}
}

@article{vae_kingma2013auto_vae,
  title={Auto-encoding variational bayes},
  author={Kingma, Diederik P and Welling, Max},
  journal={arXiv preprint arXiv:1312.6114},
  year={2013}
}

@article{vae_hwang2020variational,
  title={Variational interaction information maximization for cross-domain disentanglement},
  author={Hwang, HyeongJoo and Kim, Geon-Hyeong and Hong, Seunghoon and Kim, Kee-Eung},
  journal={Advances in Neural Information Processing Systems},
  volume={33},
  pages={22479--22491},
  year={2020}
}

@inproceedings{GAN_isola2017image,
  title={Image-to-image translation with conditional adversarial networks},
  author={Isola, Phillip and Zhu, Jun-Yan and Zhou, Tinghui and Efros, Alexei A},
  booktitle={Proceedings of the IEEE conference on computer vision and pattern recognition},
  pages={1125--1134},
  year={2017}
}

@inproceedings{GAN_wang2018high,
  title={High-resolution image synthesis and semantic manipulation with conditional gans},
  author={Wang, Ting-Chun and Liu, Ming-Yu and Zhu, Jun-Yan and Tao, Andrew and Kautz, Jan and Catanzaro, Bryan},
  booktitle={Proceedings of the IEEE conference on computer vision and pattern recognition},
  pages={8798--8807},
  year={2018}
}

@inproceedings{zhu2017unpaired,
  title={Unpaired image-to-image translation using cycle-consistent adversarial networks},
  author={Zhu, Jun-Yan and Park, Taesung and Isola, Phillip and Efros, Alexei A},
  booktitle={Proceedings of the IEEE international conference on computer vision},
  pages={2223--2232},
  year={2017}
}

@inproceedings{lee2023edge_gan,
  title={Edge-guided Multi-domain RGB-to-TIR image Translation for Training Vision Tasks with Challenging Labels},
  author={Lee, Dong--Guw and Jeon, Myung--Hwan and Cho, Younggun and Kim, Ayoung},
  booktitle={2023 IEEE International Conference on Robotics and Automation (ICRA)},
  pages={8291--8298},
  year={2023},
  organization={IEEE}
}

@article{ozkanouglu2022infragan,
  title={InfraGAN: A GAN architecture to transfer visible images to infrared domain},
  author={{\"O}zkano{\u{g}}lu, Mehmet Akif and Ozer, Sedat},
  journal={Pattern Recognition Letters},
  volume={155},
  pages={69--76},
  year={2022},
  publisher={Elsevier}
}

@article{han2024dr,
  title={DR-AVIT: Toward diverse and realistic aerial visible-to-infrared image translation},
  author={Han, Zonghao and Zhang, Shun and Su, Yuru and Chen, Xiaoning and Mei, Shaohui},
  journal={IEEE Transactions on Geoscience and Remote Sensing},
  volume={62},
  pages={1--13},
  year={2024},
  publisher={IEEE}
}

@article{sun2023vq,
  title={Vq-infratrans: A unified framework for rgb-ir translation with hybrid transformer},
  author={Sun, Qiyang and Wang, Xia and Yan, Changda and Zhang, Xin},
  journal={Remote Sensing},
  volume={15},
  number={24},
  pages={5661},
  year={2023},
  publisher={MDPI}
}

@inproceedings{T2i-adapter,
  title={T2i-adapter: Learning adapters to dig out more controllable ability for text-to-image diffusion models},
  author={Mou, Chong and Wang, Xintao and Xie, Liangbin and Wu, Yanze and Zhang, Jian and Qi, Zhongang and Shan, Ying},
  booktitle={Proceedings of the AAAI conference on artificial intelligence},
  volume={38},
  number={5},
  pages={4296--4304},
  year={2024}
}

@inproceedings{dai2025diffusion,
  title={Diffusion-based synthetic data generation for visible-infrared person re-identification},
  author={Dai, Wenbo and Lu, Lijing and Li, Zhihang},
  booktitle={Proceedings of the AAAI Conference on Artificial Intelligence},
  volume={39},
  number={11},
  pages={11185--11193},
  year={2025}
}

@article{yang2025s,
  title={S 3 OIL: Semi-Supervised SAR-to-Optical Image Translation via Multi-Scale and Cross-Set Matching},
  author={Yang, Xi and Shi, Haoyuan and Li, Ziyun and Qiao, Maoying and Gao, Fei and Wang, Nannan},
  journal={IEEE Transactions on Image Processing},
  year={2025},
  publisher={IEEE}
}

@article{chen2026any2any,
  title={Any2Any: Unified Arbitrary Modality Translation for Remote Sensing},
  author={Chen, Haoyang and Zhang, Jing and Wang, Hebaixu and Wang, Shiqin and Huang, Pohsun and Li, Jiayuan and Guo, Haonan and Wang, Di and Wang, Zheng and Du, Bo},
  journal={arXiv preprint arXiv:2603.04114},
  year={2026}
}

@article{mao2026pid,
  title={PID: Physics-Informed Diffusion Model for Infrared Image Generation},
  author={Mao, Fangyuan and Mei, Jilin and Lu, Shun and Liu, Fuyang and Chen, Liang and Zhao, Fangzhou and Hu, Yu},
  journal={Pattern Recognition},
  volume={169},
  pages={111816},
  year={2026}
}

@article{ren2024grounded,
  title={Grounded sam: Assembling open-world models for diverse visual tasks},
  author={Ren, Tianhe and Liu, Shilong and Zeng, Ailing and Lin, Jing and Li, Kunchang and Cao, He and Chen, Jiayu and Huang, Xinyu and Chen, Yukang and Yan, Feng and others},
  journal={arXiv preprint arXiv:2401.14159},
  year={2024}
}

@inproceedings{zhao2025pseudo,
  title={Pseudo-SD: pseudo controlled stable diffusion for semi-supervised and cross-domain semantic segmentation},
  author={Zhao, Dong and Zang, Qi and Wang, Shuang and Sebe, Nicu and Zhong, Zhun},
  booktitle={Proceedings of the IEEE/CVF International Conference on Computer Vision},
  pages={22393--22403},
  year={2025}
}

@inproceedings{wang2026jodiffusion,
  title={JoDiffusion: Jointly Diffusing Image with Pixel-Level Annotations for Semantic Segmentation Promotion},
  author={Wang, Haoyu and Zhang, Lei and Liu, Wenrui and Jiang, Dengyang and Wei, Wei and Ding, Chen},
  booktitle={Proceedings of the AAAI Conference on Artificial Intelligence},
  volume={40},
  number={12},
  pages={9775--9783},
  year={2026}
}

@article{Re1_Dataset_diffusion,
  title={Dataset diffusion: Diffusion-based synthetic data generation for pixel-level semantic segmentation},
  author={Nguyen, Quang and Vu, Truong and Tran, Anh and Nguyen, Khoi},
  journal={Advances in Neural Information Processing Systems},
  volume={36},
  pages={76872--76892},
  year={2023}
}

@inproceedings{Re1_joint_zhang2025paired,
  title={Paired Image Generation with Diffusion-Guided Diffusion Models},
  author={Zhang, Haoxuan and Cui, Wenju and Cao, Yuzhu and Tan, Tao and Liu, Jie and Peng, Yunsong and Zheng, Jian},
  booktitle={International Conference on Medical Image Computing and Computer-Assisted Intervention},
  pages={371--381},
  year={2025}
}

@inproceedings{semanticrt,
  title={Semanticrt: A large-scale dataset and method for robust semantic segmentation in multispectral images},
  author={Ji, Wei and Li, Jingjing and Bian, Cheng and Zhang, Zhicheng and Cheng, Li},
  booktitle={Proceedings of the 31st ACM International Conference on Multimedia},
  pages={3307--3316},
  year={2023}
}

@article{Re1_yang2023freemask,
  title={Freemask: Synthetic images with dense annotations make stronger segmentation models},
  author={Yang, Lihe and Xu, Xiaogang and Kang, Bingyi and Shi, Yinghuan and Zhao, Hengshuang},
  journal={Advances in Neural Information Processing Systems},
  volume={36},
  pages={18659--18675},
  year={2023}
}

@inproceedings{Re1_ye2024seggen,
  title={Seggen: Supercharging segmentation models with text2mask and mask2img synthesis},
  author={Ye, Hanrong and Kuen, Jason and Liu, Qing and Lin, Zhe and Price, Brian and Xu, Dan},
  booktitle={European Conference on Computer Vision},
  pages={352--370},
  year={2024},
  organization={Springer}
}

@inproceedings{sigma,
  title={Sigma: Siamese mamba network for multi-modal semantic segmentation},
  author={Wan, Zifu and Zhang, Pingping and Wang, Yuhao and Yong, Silong and Stepputtis, Simon and Sycara, Katia and Xie, Yaqi},
  booktitle={2025 IEEE/CVF Winter Conference on Applications of Computer Vision (WACV)},
  pages={1734--1744},
  year={2025},
  organization={IEEE}
}

@inproceedings{M-SpecGene,
  title={M-SpecGene: Generalized Foundation Model for RGBT Multispectral Vision},
  author={Zhou, Kailai and Yang, Fuqiang and Wang, Shixian and Wen, Bihan and Zi, Chongde and Chen, Linsen and Shen, Qiu and Cao, Xun},
  booktitle={Proceedings of the IEEE/CVF International Conference on Computer Vision},
  pages={7861--7872},
  year={2025}
}

@article{MiLNet,
  title={MiLNet: Multiplex interactive learning network for RGB-T semantic segmentation},
  author={Liu, Jinfu and Liu, Hong and Li, Xia and Ren, Jiale and Xu, Xinhua},
  journal={IEEE Transactions on Image Processing},
  year={2025},
  publisher={IEEE}
}

@inproceedings{pst900,
  title={Pst900: Rgb-thermal calibration, dataset and segmentation network},
  author={Shivakumar, Shreyas S and Rodrigues, Neil and Zhou, Alex and Miller, Ian D and Kumar, Vijay and Taylor, Camillo J},
  booktitle={2020 IEEE international conference on robotics and automation (ICRA)},
  pages={9441--9447},
  year={2020},
  organization={IEEE}
}

@inproceedings{clip,
  title={Learning transferable visual models from natural language supervision},
  author={Radford, Alec and Kim, Jong Wook and Hallacy, Chris and Ramesh, Aditya and Goh, Gabriel and Agarwal, Sandhini and Sastry, Girish and Askell, Amanda and Mishkin, Pamela and Clark, Jack and others},
  booktitle={International conference on machine learning},
  pages={8748--8763},
  year={2021},
  organization={PmLR}
}

@article{AdamW,
  title={Decoupled weight decay regularization},
  author={Loshchilov, Ilya and Hutter, Frank},
  journal={arXiv preprint arXiv:1711.05101},
  year={2017}
}

@article{zhou2017places365,
   title={Places: A 10 million Image Database for Scene Recognition},
   author={Zhou, Bolei and Lapedriza, Agata and Khosla, Aditya and Oliva, Aude and Torralba, Antonio},
   journal={IEEE Transactions on Pattern Analysis and Machine Intelligence},
   year={2017},
   publisher={IEEE}
}
}

% {
% \small

% [1] Alexander, J.A.\ \& Mozer, M.C.\ (1995) Template-based algorithms for
% connectionist rule extraction. In G.\ Tesauro, D.S.\ Touretzky and T.K.\ Leen
% (eds.), {\it Advances in Neural Information Processing Systems 7},
% pp.\ 609--616. Cambridge, MA: MIT Press.

% [2] Bower, J.M.\ \& Beeman, D.\ (1995) {\it The Book of GENESIS: Exploring
%   Realistic Neural Models with the General Neural Simulation System.}  New York:
% TELOS/Springer--Verlag.

% [3] Hasselmo, M.E., Schnell, E.\ \& Barkai, E.\ (1995) Dynamics of learning and
% recall at excitatory recurrent synapses and cholinergic modulation in rat
% hippocampal region CA3. {\it Journal of Neuroscience} {\bf 15}(7):5249-5262.
% }

%%%%%%%%%%%%%%%%%%%%%%%%%%%%%%%%%%%%%%%%%%%%%%%%%%%%%%%%%%%%

\end{document}